\documentclass[10pt,conference,letter]{IEEEtran}
\usepackage{cite}

\ifCLASSINFOpdf
\else
\fi
\usepackage{url}

\usepackage{graphics}
\usepackage{graphicx}
\usepackage{setspace} 
\usepackage{color}

\newcommand{\matrixx}{\cal{X}}
\newcommand{\vecx}{\vec{x}}
\newcommand{\vecX}{\vec{X}}
\newcommand{\mydefine}{\stackrel{def}{=}}
 
\newcommand{\melf}{\phi_{f}}
\newcommand{\linearf}{l_{f}}

\newcommand{\myalpha}{\xi}

\begin{document}
%
\title{Modified Mel Filter Bank to Compute MFCC of Subsampled Speech}

\author{\IEEEauthorblockN{Kiran Kumar Bhuvanagiri}
\IEEEauthorblockA{TCS Innovation Lab-Mumbai,
Tata Consultancy Services\\
Yantra park, Thane, Maharastra, India \\
Email: kirankumar.bhuvanagiri@tcs.com}
\and
\IEEEauthorblockN{Sunil Kumar Kopparapu}
\IEEEauthorblockA{TCS Innovation Lab-Mumbai,
Tata Consultancy Services\\
Yantra park, Thane, Maharastra, India \\
Email: sunilkumar.kopparapu@tcs.com}
}


%


\maketitle

\begin{abstract}

Mel Frequency Cepstral Coefficients (MFCCs) 
are the most popularly used speech features in
most speech and speaker recognition applications. 
In this work, we propose a modified Mel filter bank to extract MFCCs from subsampled
speech.
We also propose a stronger metric  which effectively captures the correlation between MFCCs of 
original speech and MFCC of resampled speech. 
It is found that the proposed method of filter bank construction performs distinguishably well 
and gives recognition performance on resampled speech close to recognition
accuracies on original speech. 
\end{abstract}



%
\IEEEpeerreviewmaketitle

\section{Introduction}
Time scale modification (TSM) is a class of algorithms that change the playback time of speech/audio signals. By increasing or decreasing the apparent rate of articulation, TSM on one hand, 
is useful to make degraded speech more intelligible and on the other hand, reduces the time needed for a 
listener to listen to a message. Reducing the playback time of speech or {\em
time compression of speech signal} has a variety of applications 
that include teaching aids to the disabled and in human-computer interfaces. 
Time-compressed speech is also referred to as accelerated, compressed, time-scale modified, sped-up, 
rate-converted, or time-altered speech. 
Studies have indicated that listening to teaching materials twice that have been speeded up by a 
factor of two is more effective than listening to them once at normal speed \cite{barry92}. 
Time compression techniques have also been used in speech recognition systems to time 
normalize input utterances to a standard length. One potential application is that TSM is often used to 
adjust Radio commercials and the audio of television advertisements to fit
exactly into the $30$ or $60$ seconds. 
Time compression of speech also saves 
storage space and transmission bandwidth for speech messages. 
Time compressed speech has been used to speed up message presentation in voice
mail systems \cite{hejn90}. 

In general, time scale modification of a speech signal 
is associated with a parameter called time scale modification (TSM) 
factor or scaling factor. 
In this paper we denote the TSM factor by $\alpha$. 
There are a variety of techniques for time scaling of speech out of which, 
resampling is one of the simplest techniques. 
Resampling of digital signals is basically a process of decimation (for time compression, $\alpha>1$) or 
interpolation (for time expansion, $\alpha<1$) or a combination of both. 
Usually, for decimation, the input signal is 
subsampled. 
For interpolation, {\em zeros} are inserted between samples of the original 
input signal.
For a discrete time signal $x[n]$ the restriction on the TSM factor $\alpha$ to
obtain $x[\alpha n]$ is
that $\alpha$  be a rational number. 
For any $\alpha = \frac{p}{q}$ where $p$ 
and $q$ are integers the signal $x[\alpha n]$ is constructed by 
first interpolating $x[n]$ by a factor of $p$, say $x^p = x[n \uparrow p]$ and then
decimating $x[n]$ by a factor of $q$, namely, $x^q = x[n \downarrow q]$. 
It should be noted that, usually interpolation is carried out before decimation
to eliminate information loss in the pre-filtering of decimation. 

Most often, cepstral features are the 
speech features of choice for many speaker and speech recognition 
systems. For example, the Mel-frequency cepstral coefficient (MFCC) 
\cite{merm80} representation of speech is probably the most commonly 
used representation in speaker recognition and and speech recognition applications
\cite{reyn95, 
rash04, sedd04}. In general, cepstral features are 
more compact, discriminable, and most importantly, nearly 
decorrelated such that they allow the diagonal covariance to be used by 
the hidden Markov models (HMMs) effectively. Therefore, they can usually 
provide higher baseline performance over filter bank features 
\cite{zhan04}. 

In \cite{sunil2010}  
6 types of filter banks were proposed to calculate MFCC's on subsampled speech
and used Pearson coefficient as a measure of similarity to compare the MFCC of
subsampled speech and the MFCC of original speech. In the present work, 
a Mel filter bank construct is presented that is able to 
extract significantly more correlated MFCC's of the
subsampled speech with respect to the MFCC's of original speech.
We also experimentally show that the new  Mel filter band
construction 
performs better than all the six approaches  mentioned in \cite{sunil2010} in
addition we
perform
speech recognition experiments on AN4 speech database \cite{link1} 
using 
open source ASR engine \cite{link2} to show the recognition performance on
resampled speech is as good as the recognition accuracy on original speech.
One of the application of this work is in the  scenario where training is 
done on original speech and decoding has to be done on 
 subsampled speech. 
In Section \ref{sec:mfcc_theory} procedure to compute MFCC features is 
discussed. In Section \ref{sec:theory} 
we derive a relationship between the MFCC parameters computed for original
 speech
and the time scaled speech. 
In Section \ref{sec:proposed}, new filter bank approach is proposed.  
Section \ref{sec:expts} 
gives the details of the experiments conducted to substantiate advantage of
proposed modified filter bank and
we conclude in Section \ref{sec:conclusions}.


\section{Computing the MFCC parameters \cite{sunil2010}}
\label{sec:mfcc_theory}
\begin{figure}
\centering
\includegraphics[width=0.45\textwidth]{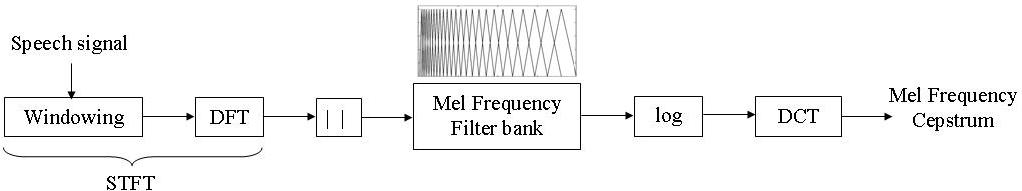}
\caption{Computation of Mel Frequency Cepstral Coefficients \cite{sunil2010}}
\label{fig:mfcc1}
\end{figure}
The outline of the computation of Mel frequency cepstral coefficients (MFCC)
 is shown in Figure \ref{fig:mfcc1}. 
In general, the MFCCs are computed as follows. 
Let $x[n]$ be a speech signal with a sampling frequency of $f_s$, and is divided into $P$ 
frames each of length $N$ samples with an overlap of $N/2$ samples such that 
$\left \{ \vecx_{1}[n],  
\vecx_{2}[n] \cdots \vecx_{p}[n] \cdots \vecx_{P}[n] \right \} $, 
where $\vecx_{p}[n]$ denotes the $p^{th}$ frame of the speech signal
$x[n]$ and is
$\vecx_p[n] = \left \{ x \left [p* \left (\frac{N}{2}-1 \right )+i \right ]  \right \}_{i=0}^{N-1}$
Now the speech signal $x[n]$ can be represented 
in matrix notation as 
$ {\matrixx} \mydefine  
[\vecx_{1}, \vecx_{2}, \cdots, \vecx_{p}, \cdots, \vecx_{P}] $. 
Note that the size of the matrix $\matrixx$ is $N \times P$. 
The MFCC features
are computed for each frame of the speech sample (namely, for all $\vecx_{p}$).

In speech signal processing, in order to compute the MFCCs of 
the $p^{th}$ frame, $\vecx_{p}$ is multiplied with a hamming window 
$w[n] = 0.54 - 0.46 \cos \left ( \frac{n \pi}{N} \right )$, followed by the
discrete Fourier transform (DFT) as shown in (\ref{eq:dft}) \cite{open89}.
\begin{equation} 
X_{p}(k) = \sum_{n=0}^{N-1} x_p[n] w[n]\exp^{-j \frac{2\pi kn}{N}} 
\label{eq:dft} 
\end{equation} 
for $k=0, 1, \cdots, N-1$. 
If $f_{s}$ is the sampling rate of the speech signal $x[n]$ then
$k$ 
corresponds to the frequency $\linearf(k) = kf_{s}/N$.
Let $\vecX_{p} = [X_{p}(0), X_p(1),  \cdots,  X_{p}(N-1)]^T$ represent the 
DFT of the windowed $p^{th}$ frame of the speech signal $x[n]$, namely $\vecx_p$.
Accordingly, let $X = [\vecX_{1}, \vecX_{2}, \cdots \vecX_{p}, \cdots, \vecX_{P}]$ represent the DFT of the 
matrix $\matrixx$.
Note that the size of $X$ is $N \times P$ and is known as  STFT (short time
Fourier transform) matrix.
The modulus of Fourier transform is extracted and the magnitude spectrum 
is obtained as $|X|$ which again is a matrix of size $N$ x $P$. 

The modulus of Fourier transform is extracted and the magnitude spectrum 
is obtained as $|X|$ which is a matrix of size $N \times P$. 
The magnitude spectrum is warped according to the Mel scale in order to 
adapt the frequency resolution to the properties of the human 
ear \cite{sirk01}. 
Note that the Mel ($\melf$) and the linear frequency ($\linearf$) \cite{thom04} are
related, namely, 
$\melf = 2595 * log_{10}(1+\frac{\linearf}{700})$ 
where $\melf$ is the Mel frequency and $\linearf$ is the linear frequency. 
Then the magnitude spectrum $|X|$ 
is segmented into a number of critical bands by means of a 
Mel filter bank which typically consists of a series of overlapping 
triangular filters defined by their center frequencies 
$\linearf{_{c}}(m)$. 

The parameters that define a Mel filter bank are 
(a) number of Mel filters, $F$, (b) minimum frequency, $\linearf{_{min}}$ and 
(c) maximum frequency, $\linearf{_{max}}$. 
For speech, in general, it is suggested in \cite{link} 
that $\linearf{_{min}} > 100$ Hz. 
Furthermore, by setting $\linearf{_{min}}$ above 50/60Hz, we get rid of the 
hum resulting from the AC power, if present. 
\cite{link} also suggests that $\linearf{_{max}}$ be less than the 
Nyquist frequency. Furthermore, there is not much information above $6800$ Hz. 
Then a fixed frequency resolution in the Mel scale is computed using 
$\delta \melf = (\melf{_{max}}-\melf{_{min}})/(F+1)$ 
where $\melf{_{max}}$ and $\melf{_{min}}$ are the frequencies on the Mel scale corresponding 
to the linear frequencies $\linearf{_{max}}$ and $\linearf{_{min}}$ 
respectively. 
The center frequencies on the Mel scale are given by $\melf{_c(m)} = m.\delta
\phi$ where $m=1, 2, \cdots, F$.
To obtain the center frequencies of the triangular Mel filter bank in Hertz, we use the inverse relationship 
between $\linearf$ and $\melf$ given by $\linearf{_c(m)} =
700(10^{\melf{_c(m)}/2595}-1)$. The Mel filter bank, $M(m,k)$ \cite{pr_sigu06} is given by 
\[
\mbox{\scriptsize{$M(m,k)$}}=\left\{
\begin{array}{ll}
\mbox{\scriptsize{0}} & \mbox{\scriptsize{for $\linearf(k)<\linearf{_{c}(m-1)}$}} \\
\mbox{\scriptsize{$\frac{\linearf(k)-\linearf{_c(m-1)}}{\linearf{_{c}(m)}-\linearf{_{c}(m-1)}}$}}
& \mbox{\scriptsize{for $\linearf{_{c}(m-1)}\le \linearf(k) < \linearf{_{c}(m)}$}} \\
\mbox{\scriptsize{$\frac{\linearf(k)-\linearf{_c(m+1)}}{\linearf{_{c}(m)}-\linearf{_{c}(m+1)}}$}} & 
\mbox{\scriptsize{for $\linearf{_{c}(m)}\le \linearf(k) < \linearf{_{c}(m+1)}$}} \\
\mbox{\scriptsize{0}} & \mbox{\scriptsize{for $ \linearf(k)\ge
\linearf{_{c}(m+1)}$}}
\end{array} \right .
\]
The Mel filter bank $M(m,k)$ is an $F \times N$ matrix. 

The logarithm of the filter bank outputs (Mel spectrum) is given in (\ref{ln}).
\begin{equation}
L_p(m,k) = ln\left \{\sum_{k=0}^{N-1}M(m,k) * |X_p(k)| \right \}
\label{ln}
\end{equation}
where $m = 1, 2, \cdots,  F$ and $p = 1, 2, \cdots, P$. 
The filter bank output, which is the product of the Mel filter bank, 
$M$ and the magnitude spectrum, $|X|$ is a $F \times P$ matrix. 
A discrete cosine transform of $L_p(m,k)$ results in the MFCC parameters. 
\begin{equation}
\label{dct}
\Phi_p^r \left \{ x[n] \right \} = \sum_{m=1}^{F}L_p(m,k) \cos \left \{\frac{r(2m-1)\pi}{2F} \right \}
\end{equation}
where $r = 1, 2, \cdots,  F$ and $\Phi_p^r \left \{ x[n] \right \}$ represents
the $r^{th}$ MFCC of the $p^{th}$ frame of the speech signal $x[n]$. 
The MFCC of all the $P$ frames of the speech signal are obtained as a matrix 
$\Phi$ 
\begin{equation}
\Phi \left \{\cal{X} \right \} = [\Phi_{1},  \Phi_{2}, \cdots,  \Phi_{p}, \cdots \Phi_{P}]
\end{equation}
Note that the $p^{th}$ column of the matrix $\Phi$, namely $\Phi_p$ represents the MFCC of the 
speech  signal, $x[n]$, corresponding to the $p^{th}$ frame, $x_p[n]$.

\section{MFCC of Resampled Speech \cite{sunil2010}}
\label{sec:theory}
In this section, we show how the resampling of the speech signal 
in time effects the computation of MFCC parameters. 
Let $y[s]$ denote the time scaled speech signal given by 
\begin{equation}
y[s] = x[\alpha n]  = x \downarrow \alpha
\end{equation} 
where $\alpha$ is the scaling
factor.
Let $y_{p}[s] = x_{p}[\alpha n]  = x_p \downarrow \alpha$ 
denote the $p^{th}$ frame of the time scaled speech where $s=0, 1,\cdots, S-1$, 
$S$ being the number of samples in the time scaled speech frame 
given by $S=\frac{N}{\alpha}$. 
If $\alpha < 1$ the signal is expanded in time while
$\alpha > 1$ means the signal is compressed in time. 
Note that if $\alpha = 1$ the signal remains unchanged. 
DFT of the windowed $y_p[n]$ is calculated from the DFT of $x_p[n]$.
Assuming that $\alpha$ is an integer and using the scaling property of DFT, 
 we have,
\begin{equation}
Y_{p}(k') = \frac{1}{\alpha}\sum_{l=0}^{\alpha-1}X_{p}(k' + lS )
\label{eq:sampled_spectrum}
\end{equation}
where $k' = 1, 2, \cdots, S$.
The MFCC of the time scaled speech are given by
\begin{equation}
\label{eq:mfcctsm}
\Phi_p^r\{y[n]\} =  \Phi_{p}^{r} \left \{ x\downarrow\alpha \right \} 
= \sum_{m=1}^{F} L_p'(m,k') cos \left \{\frac{r(2m-1)\pi}{2F} \right \}
\end{equation}
where $r = 1, 2, \cdots, F$ and
\begin{equation}
L_p'(m,k') = ln \left \{\sum_{k'=0}^{S-1}{M'(m,k')}  \left |\frac{1}{\alpha}\sum_{l=0}^{\alpha-1}X_{p}(k' + lS)\right | \right \}
\label{eq:lns}
\end{equation}
Note that $L_p'$ and $M'$ are the log Mel spectrum and the Mel filter bank of the 
resampled speech. 
We propose a filter bank, $M'(m,k')$ which 
is used in the calculation of MFCC 
of the resampled speech. Note that a good choice of the Mel filter bank is the one 
which gives (a) the best Pearson correlation between the 
MFCC of the original speech and the MFCC of the resampled speech and (b) 
best speech recognition accuracies when trained using the original speech and
decoded using the subsamples speech. 

\section{Modified Filter Bank}
\label{sec:proposed}

One of the major steps in the computation of MFCC of 
subsampled speech is through construction of the
Mel filter bank. 
%
The proposed Mel filter bank $M_{new}(m,k')$ for subsampled speech is given as
\[
\mbox{{$M_{new}(m,k')$}}=\left\{
\begin{array}{ll}
\mbox{\scriptsize{$M(m,\alpha k')$}} & \mbox{\scriptsize{for
$\linearf(k') \le (\frac{1}{\alpha} \frac{f_s}{2} )$}} \\
\mbox{\scriptsize{0}}
& \mbox{\scriptsize{for $\linearf(k')>(\frac{1}{\alpha} \frac{f_s}{2})$}} \\
\end{array} \right.
\]
where $k'$ ranges from $1$ to $N/2$. 
Notice that the modified filter bank is the subsampled version of original
filter bank with bands above $f_s/2\alpha$ set to $0$. 
In other words the center frequencies and bandwidths  of proposed filter bank and
original filter bank are the same for $f_c < f_s/2\alpha$. 
In order to keep the total number of filter bank outputs same as original 
we fill in the remaining filter bank outputs in the following manner.
Let $F_{\myalpha}$ be the number of filter banks whose $f_c$ is below
$f_s/2\alpha$ and
let $L_{new,p}(m,k')$ be output of the $p^{th}$ Mel filter bank. 
Then we define outputs of
Mel filter banks 
\begin{equation}
L_{new,p}(m,k')=(0.95)^{m-F_{\myalpha}}L_{new,p}(F_{\myalpha},k')
\label{eq:fillinmops}
\end{equation}
where $F_{\myalpha}< m \le F$. 

In all our experiments, we assume, 
(a) $\alpha=2$, 
(b) the number of Mel filters used for the feature extraction of 
original speech and that of the resampled speech are same and, 
(c) the window length reduces by half, namely, $N/2$. 
 
 \section{Experimental Results}
 \label{sec:expts}

We conducted experiments on AN4 audio database. 
It consists of $948$ 
train  and $130$ test  audio files and the test audio files contain $773$
spoken words or phrases. The recognition results are based on these $773$ words
and phrases.
All the speech files in the AN4 database are sampled at $16$ KHz. 
The Mel filter bank used has $F = 30$ bands spread from $\linearf{_{min}} = 130$ Hz 
to a maximum frequency of  $\linearf{_{max}} = 6800$ Hz and the frame size if
of $32$ ms.
The MFCC parameters 
(denoted by $\Phi \{x[n]\} = 
[\Phi_{1},  \Phi_{2}, \cdots,  \Phi_{m}, \cdots \Phi_{F}]$\footnote{Note that $\Phi_m$ is a vector formed with the $m^{th}$ MFCC of all the speech frames}) are computed for 
the $16$ kHz speech signal $x[n]$, as described in 
Section \ref{sec:mfcc_theory}. 
We then subsampled $x[n]$ by a factor of $\alpha = 2$ and 
constructed $y[s] = x \downarrow 2 = x[2n]$. 
The MFCC parameters of $y[s]$ (denoted by $\Phi \{y[s]\} = 
[\Phi'_{1},  \Phi'_{2}, \cdots,  \Phi'_{m}, \cdots \Phi'_{F}]$) 
are calculated using the proposed Mel filter bank (\ref{eq:fillinmops}). 
We conducted two types of experiments to evaluate the performance of using the
proposed Mel filter bank construction. 
We used, as done in \cite{sunil2010}, Pearson coefficient to compare the MFCC
of the subsampled speech with the MFCC of the original speech; in addition we
used  speech recognition accuracies to compute the appropriateness of the
Mel filter bank construction. Pearson correlation coefficient (denoted by $r$)
is computed between the MFCC parameters of the subsampled speech (using
different Mel-filter bank constructs) and the MFCC of the original speech  
in two different ways  and the 
speech recognition experiments were done using Sphinx ASR. 
We trained HMM models using train data set using the Sphinx toolbox and
accuracies were calculated on the test data using Sphinx.

\subsection{Comparison using Pearson Coefficient}
We considered two types of evaluations. In  {\em Case I}
the $F$-dimensional MFCC vector of each frame is concatenated and the $r$ between the 
MFCC of the original speech and the MFCC of the subsampled speech  is computed.
The mean and variance of Pearson correlation coefficients, $r$ for the  proposed method and $6$ methods of \cite{sunil2010}
are shown in Table \ref{tab:pc_score} for $130$ test samples. 
\begin{table}
\begin{center}
\caption{Pearson coefficient ($r$) between the MFCC of original and subsampled speech}
\begin{tabular}{|c|c|c||c|c|}
\hline
 Filter bank & \multicolumn{2}{c||}{{\em Case I}}  &\multicolumn{2}{c|}{{\em
Case II}} \\ \cline{2-5} 
Type & mean & variance & mean & variance \\
\hline  
 {Proposed} & {\bf 0.986} & 0.00004 & {\bf 0.961} & 0.00433\\ 
 {A} \cite{sunil2010} & 0.868 & 0.00196 & 0.692 & 0.14715\\ 
 {B} \cite{sunil2010} & {\bf 0.913} & 0.00087 & {\bf 0.756} &  0.07570\\
 {C} \cite{sunil2010}& 0.906 & 0.00104 & 0.746 &  0.08460\\
 {D} \cite{sunil2010}& 0.842 & 0.00266 & 0.662 &  0.16834\\
 {E} \cite{sunil2010}& 0.756 & 0.00572 & 0.619 & 0.15979\\
 {F} \cite{sunil2010}& 0.798 & 0.00431 & 0.680 & 0.10893\\
\hline
\end{tabular}
\label{tab:pc_score} 
\end{center}
\end{table}
In {\em Case II} 
instead of concatenating MFCCs of all the frames, 
we computed $r$ for MFCCs of each frame of subsampled speech and the original speech. 
The mean and variances of $r$ are shown in Table \ref{tab:pc_score}. Clearly
the proposed Mel filter bank construction performs better than the ones
suggested in \cite{sunil2010} in both the cases. Compare $0.986$ of the proposed
method to the $0.913$ best of \cite{sunil2010} for {\em Case I} and $0.961$
compared to $0.756$ for {\em Case II}.

\subsection{Speech recognition Performance}
\begin{table}
\begin{center}
\caption{Recognition accuracies (percentage)}
\begin{tabular}{|c|c|c|} \hline
 Filter bank Type & {\em Case A} & {\em Case B} \\ \hline  
{Original ($16$ kHz)} & {\bf 43.21} & {\bf 81.63} \\ \hline
{Proposed ($8$ kHz)} & {\bf 37.00} & {\bf 77.62} \\
 {A ($8$ kHz) \cite{sunil2010}} & 1.42 & 3.62 \\
 {B ($8$ kHz) \cite{sunil2010}} & 2.72 &  11.00\\
 {C ($8$ kHz) \cite{sunil2010}} & {\bf 3.88} &  {\bf 11.90}\\
 {D ($8$ kHz) \cite{sunil2010}} & 0.78 &  2.85\\
 {E ($8$ kHz) \cite{sunil2010}} & 1.68 & 1.03 \\
 {F ($8$ kHz) \cite{sunil2010}} & 1.94 & 2.46 \\ \hline
\end{tabular}
\label{tab:Decodeoutput} 
\end{center}
\end{table}

\begin{figure} 
\centering
\includegraphics[width=0.45\textwidth]{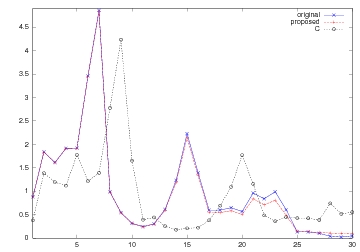}
\caption{Sample Filter bank outputs of original speech, and subsampled speech
using the proposed Mel filter bank and the best Mel filter bank in
\cite{sunil2010}}
\label{fig:melops}
\end{figure}

We used $948$ training speech samples of AN4 database to build acoustic models
using Sphinx train tool box. Training is done using speech features calculated
on the $16$ kHz (original) speech files. 
Recognition is performed on the $130$ test speech samples, both for the
original ($16$ kHz) and subsampled ($8$ kHz) speech. 
In {\em Case A} we extracted $30$ MFCC's while in {\em Case B} 
we extracted $30$ MFCC and used only the first $13$ of them and appended them with
$13$ velocity and $13$ acceleration coefficients to form a $39$ dimensional feature vector. 
Recognition accuracies are  shown in Table \ref{tab:Decodeoutput} as word
recognition
rate on the $773$ words in the $130$ test speech files.
It can be observed that word  recognition accuracies 
using the proposed Mel filter bank on subsampled speech is closer to  the baseline word
recognition accuracies calculated on the
original speech for both {\em Case A} and {\em Case B} and much higher than the
best Mel filter bank in \cite{sunil2010}. Compare $37$\% to $3.88$\% and
$77.62$\% to
$11.90$\% in Table \ref{tab:Decodeoutput} for {\em Case A} and {\em Case B}
respectively.
The improved performance of the proposed Mel filter bank in terms of recognition
accuracies can be explained by looking at a sample filter bank output shown 
in Fig.\ref{fig:melops}. 
Filter bank output of the proposed Mel filter bank construct (red line; '+')
closely follows that of the
original speech Mel filter bank output (blue line; 'x'), while that of C (best
in \cite{sunil2010}) filter
bank (black line; 'o') 
shows a shift in the filter bank outputs. 
This is primarily because the center frequencies and bandwidths of filter bank
(C in \cite{sunil2010}) are different from original filter bank, 
but this is not true in the method proposed in this paper.

\section{Conclusion}
\label{sec:conclusions}
We proposed a modified Mel filter bank  to extract MFCC from
subsampled speech which correlates well with the MFCC of the original speech.
We showed that the proposed Mel filter bank outperforms all the Mel filter
banks developed in \cite{sunil2010} experimentally both in terms of correlation
with the MFCC of the original speech and also through word recognition
accuracies. The primary importance of this work is when there are available 
trained models for speech of one sampling frequency and the recognition has to
be performed at a subsampled or compressed speech. The suggested
Mel filter bank makes it possible to use the HMM models of the original
speech without having to train the speech engine {\em again} with the subsampled
speech.



%



\bibliographystyle{IEEEtran}
\bibliography{TSM_MFCC}

\begin{thebibliography}{10}
\providecommand{\url}[1]{#1}
\csname url@samestyle\endcsname
\providecommand{\newblock}{\relax}
\providecommand{\bibinfo}[2]{#2}
\providecommand{\BIBentrySTDinterwordspacing}{\spaceskip=0pt\relax}
\providecommand{\BIBentryALTinterwordstretchfactor}{4}
\providecommand{\BIBentryALTinterwordspacing}{\spaceskip=\fontdimen2\font plus
\BIBentryALTinterwordstretchfactor\fontdimen3\font minus
  \fontdimen4\font\relax}
\providecommand{\BIBforeignlanguage}[2]{{%
\expandafter\ifx\csname l@#1\endcsname\relax
\typeout{** WARNING: IEEEtran.bst: No hyphenation pattern has been}%
\typeout{** loaded for the language `#1'. Using the pattern for}%
\typeout{** the default language instead.}%
\else
\language=\csname l@#1\endcsname
\fi
#2}}
\providecommand{\BIBdecl}{\relax}
\BIBdecl

\bibitem{barry92}
B.~Arons, ``Techniques, perception, and applications of time- compressed
  speech,'' \emph{Proceedings of American Voice I/O Society}, pp. 169--177,
  Sep. 1992.

\bibitem{hejn90}
D.~J. Hejna, ``Real-time time-scale modification of speech via the synchronized
  overlap-add algorithm,'' Master's thesis, MIT, Department of Electrical
  Engineering and Computer Science, 1990.

\bibitem{merm80}
S.~B. Davis and P.~Mermelstein, ``Comparison of parametric representations for
  monosyllabic word recognition in continuously spoken sentences,'' \emph{IEEE
  Trans. Acoust. Speech Signal Processing}, vol.~28, no.~4, pp. 357--366, 1980.

\bibitem{reyn95}
D.~A. Reynolds and R.~C. Rose, ``Robust text-independent speaker identification
  using {G}aussian mixture speaker models,'' \emph{IEEE Transactions on Speech
  and Audio Processing}, vol. 3, No. 1, January 1995.

\bibitem{rash04}
M.~R. Hasan, M.~Jamil, M.~G. Rabbani, and M.~S. Rahman, ``Speaker
  identification using {M}el frequency cepstral coefficients,'' \emph{3rd
  International Conference on Electrical \& Computer Engineering ICECE 2004},
  28-30 December 2004, Dhaka, Bangladesh.

\bibitem{sedd04}
H.~Seddik, A.~Rahmouni, and M.~Sayadi, ``Text independent speaker recognition
  using the {M}el frequency cepstral coefficients and a neural network
  classifier,'' \emph{First International Symposium on Control, Communications
  and Signal Processing}, pp. 631--634, 2004.

\bibitem{zhan04}
Z.~Jun, S.~Kwong, W.~Gang, and Q.~Hong, ``Using {M}el-frequency cepstral
  coefficients in missing data technique,'' \emph{EURASIP Journal on Applied
  Signal Processing}, vol. 2004, no. 3, pp. 340--346, 2004.

\bibitem{sunil2010}
S.~Kopparapu and L.~Narayana, ``Choice of {M}el filter bank in computing {MFCC}
  of a resampled speech,'' \emph{International Conference on Information
  Science, Signal processing and their applications (ISSPA)}, 2010.

\bibitem{link1}
\BIBentryALTinterwordspacing
CMU. {AN4} database. [Online]. Available:
  \url{http://www.speech.cs.cmu.edu/databases/an4/}
\BIBentrySTDinterwordspacing

\bibitem{link2}
\BIBentryALTinterwordspacing
------. Sphinx. [Online]. Available: \url{http://www.speech.cs.cmu.edu/}
\BIBentrySTDinterwordspacing

\bibitem{open89}
Oppenheim and Schafer, \emph{Discrete Time Signal Processing}.\hskip 1em plus
  0.5em minus 0.4em\relax Prentice-Hall, 1989.

\bibitem{sirk01}
S.~Molau, M.~Pitz, R.~S. Uter, and H.~Ney, ``Computing {M}el-frequency cepstral
  coefficients on the power spectrum,'' \emph{Proc. Int. Conf. on Acoustic,
  Speech and Signal Processing}, pp. 73 -- 76, 2001.

\bibitem{thom04}
T.~F. Quatieri, ``Discrete-time speech signal processing: Principles and
  practice,'' \emph{Pearson Education}, vol.~II, pp. 686, 713, 1989.

\bibitem{link}
\BIBentryALTinterwordspacing
CMU. Mel filter bank. [Online]. Available:
  \url{http://cmusphinx.sourceforge.net/sphinx4/javadoc/edu/cmu/sphinx/fronten%
d/frequencywarp/MelFrequencyFilterBank.html}
\BIBentrySTDinterwordspacing

\bibitem{pr_sigu06}
S.~Sigurdsson, K.~B. Petersen, and T.~L. Schiøler, ``{M}el frequency cepstral
  coefficients: An evaluation of robustness of {MP3} encoded music,''
  \emph{Conference Proceedings of the Seventh International Conference on Music
  Information Retrieval (ISMIR)}, Vicoria, Canada, 2006.

\end{thebibliography}

\end{document}